\documentclass[12pt]{article}
\usepackage{graphicx}
\usepackage{amsfonts}
\usepackage{latexsym}
\usepackage{amsmath}

\makeatletter
\@addtoreset{figure}{section}
\def\thefigure{\thesection.\@arabic\c@figure}
\@addtoreset{table}{section}
\def\thetable{\thesection.\@arabic\c@table}

\def\@sect#1#2#3#4#5#6[#7]#8{\ifnum #2>\c@secnumdepth
     \def\@svsec{}\else
     \refstepcounter{#1}\edef\@svsec{\csname the#1\endcsname.\hskip .75em
}\fi
     \@tempskipa #5\relax
      \ifdim \@tempskipa>\z@
        \begingroup #6\relax
          \@hangfrom{\hskip #3\relax\@svsec}{\interlinepenalty \@M #8\par}%
        \endgroup
       \csname #1mark\endcsname{#7}\addcontentsline
         {toc}{#1}{\ifnum #2>\c@secnumdepth \else
                      \protect\numberline{\csname the#1\endcsname}\fi
                    #7}\else
        \def\@svsechd{#6\hskip #3\@svsec #8\csname #1mark\endcsname
                      {#7}\addcontentsline
                           {toc}{#1}{\ifnum #2>\c@secnumdepth \else
                             \protect\numberline{\csname the#1\endcsname}\fi
                       #7}}\fi
     \@xsect{#5}}
\def\@begintheorem#1#2{\it \trivlist \item[\hskip \labelsep{\bf #1\ #2.}]}
\def\section{\@startsection {section}{1}{\z@}{-3.5ex plus -1ex minus
 -.2ex}{2.3ex plus .2ex}{\normalsize\bf}}

\begin{document}

\title{Why Two Sexes?}
\date{} 
\maketitle

\begin{center}
\author{Vigen A. Geodakian}
\end{center}

\setlength{\baselineskip}{0.995\baselineskip}
\normalsize
\vspace{0.5\baselineskip}
\vspace{1.5\baselineskip}

\begin{abstract}
Evolutionary role of the separation into two sexes 
from a cyberneticist's point of view.
[I translated this 1965 article from Russian 
"Nauka i Zhizn" (Science and Life) in 1988. 
In a popular form, the article puts forward 
several useful ideas not all of which even today are necessarily 
well known or widely accepted. {\em Boris Lubachevsky, bdl@bell-labs.com} ]
\end{abstract}
\section{Are males really needed?}
\hspace*{\parindent}
Interesting lizards live on the mountains of Armenia,
by the shore of Lake Sevan.
A peculiarity of the species is that they do not have males;
their females deposit non-fertilized
eggs from which only females hatch.
   
Such a reproductive method is extremely simple
and very rational: 
each animal can produce progeny and the
difficulties of finding a "spouse"
are eliminated.
It turns out that
the task of reproduction
can be performed quite well without males.
   
Another interesting method of reproduction
is demonstrated by 
silver crucian carps {\em carassius auratus},
inhabitants of Russian
lakes.
Like the Lake Sevan lizards,
only females represent the species.
These females do resort to the service
of males, but  .... of different fish species.
Sperm of "foreign" males stimulates
the roe to develop.
However, real fertilization,
i.e. the fusion of nucleus of male and female
sex cells, does not occur.
The males 
do not generate 
new organisms genetically
and can not claim the fatherhood.
A needle or certain chemicals
can play the role of a father.
For example, the frog roe 
can be stimulated to develop
by the prick of a thin needle,
and cell division in the eggs 
of certain sea species
is triggered
by shaking or adding 
certain acids or salts to the water.
  
In the laboratory,
even such a highly organized animal as a rabbit
can be born without a father.
The experimenters sometimes succeed 
in triggering cell division mechanically or chemically 
in an ovum extracted from a female rabbit;
the result is then
placed back into the womb of the mother.
Later, the mother delivers 
a normal live newborn "orphan" rabbit,
which can become a normal adult animal.
[This method should not be confused with the 
much publicized {\em in vitro} fertilization on humans
where the biological father, possibly anonymous, always exists.
{\em Translator's comment}.]
  
In some species the attitude toward the male is very "unfair."
For example, a spider female allows the male to copulate
but eats it up immediately after the "marriage."
To avoid this destiny the male must fetch some tasty food
to its bloodthirsty bride.
Despite its name,
the female of
the praying mantis
{\em mantis religiosa}
behaves "godlessly."
During copulation, it bites away
the head of the male so that
the latter has to complete its
mission with no head.
  
This set of examples
can be augmented by the
habits of bees
which "permit"
the males to be born only in some
generations, or practice
their elimination immediately
after the female is fertilized.
  
However, in the majority
of species the females "keep"
their males, tolerate them
and treat them fairly well.
Moreover, some species which
"know" how to reproduce without
males, e.g. some crayfish,
manage without the males in summer,
when it is warm and there is enough food.
But as soon as the fall or
a drought begins they resort 
to the service of males.
This makes one believe that
there is still a purpose in males.
  
Let us try to figure out this purpose.

\section{Why is the crossbreeding needed?}
\hspace*{\parindent}
Two main methods of reproduction exist:
asexual and sexual.
Only one parent organism participates
in the asexual method, producing
organisms similar to itself.
Two parents participate
in sexual reproduction.
However, the principal feature is not the
quantitative one:
"two from one" in the first case
and
"three from two" in the second case.
Much more important is the qualitative feature:
in the asexual method no new quality appears,
whereas in each instance of the sexual reproduction
new qualities appear which are different from the parental ones.
  
The asexual method is encountered mostly in one-cell species,
whereas most animals including mammals
reproduce sexually.
This suggests that the latter method is more
progressive from the viewpoint of evolution.
The evolutionary advantages of sexual reproduction
usually
are attributed to the recombination of genes
leading to a genetic variety.
  
Sexual reproduction necessarily assumes crossbreeding,
and, as a rule, is accompanied by splitting
into two sexes.
It is the crossbreeding which 
generates new gene combinations.
But why are two {\em different} sexes needed?

\section{Why are two sexes needed?}
\hspace*{\parindent}
A method of reproduction exists in which
the animals are not separated -
not differentiated - into two sexes
but crossbreeding still takes place.
This method is used by earthworms.
Each worm is both the male and the female.
Oysters behave similarly:
each animal first behaves as a male
and then as a female.
  
This method seems to have many advantages.
Indeed, suppose a population consists of
100 animals and each of them without exception can
breed with all the other.
Then the number of possible combinations
is $100 \times 99 / 2 ~=~ 4950$.
However, if the same population is split fifty-fifty 
into two sexes,
then the number of possible combinations
is about half as large: $50 \times 50 ~=~2500$.
The division into sexes makes things worse, as it seems,
because
a sexually divided population 
sacrifices about half of the possible genetic combinations in each generation
compared with a non-divided one.
What advantages are
received by the sexually divided population in return?
  
A common belief is 
that the advantage is the differentiation
which provides two sorts of gametes, i.e., sex cells:
small mobile spermatozoa,
whose task is to reach the ova,
and
relatively big but immobile ova,
which provide the future embryos
with nutrients.
However, a similar specialization takes place
among the hermaphroditic animals
(e.g., earthworms and oysters)
without sex differentiation 
and the accompanied decrease
in the genetic variety.
Hence we can not explain the biological meaning
of sex differentiation in this way.

Let us try to analyze the role of the sexes
in the reproduction process,
i.e., to elucidate
their relationships to
the main production criteria:
quantity,
quality, and variety of the product.
Because each kind of mass-production is
mainly characterized by these three parameters.

\section{Quantity, quality, and variety of the product}
\hspace*{\parindent}
Suppose 100 bisons are released into a sanctuary.
How should the ratio between the sexes
be chosen,
how many cows and how many bulls?
Obviously, this ratio depends
on the goal.
For example, if the goal is to maximize
the {\em quantity},
the number of calves,
then 99 cows and 1 bull is a reasonable proportion,
because 99 new calves can be born in each generation.
However these calves will all share the same father,
and will differ only as to the mother.
The number of possible parent combinations in this case is 99.
  
If, on the other hand, maximum
{\em variety}
of the progeny
is desired,
then the number of cows should equal the number of bulls.
In this case, 
the number of possible combinations 
is $50 \times 50 ~=~ 2500$.
However, the number of progeny decreases,
because only 50 calves will be born in each generation.
  
Finally if the {\em quality} of the herd is the goal,
then conditions for sexual selection
should be created 
in such a way
that part of the animals do not participate
in the reproduction.
For this it is necessary to have extra males.
Then,
the competition for the females will eliminate
the representation of some of the males from the progeny.
The larger the ratio of males to females,
the more severe is the selection.
  
Thus, there exists a kind of specialization of function
between the sexes in reproduction.
The two sexes have different relationships
with respect to the main parameters,
quantity and quality of the progeny:
the more females in the population, the higher
the quantity, 
while the more males in the population, 
the better the selection,
and the faster the changes in quality.
  
This asymmetry appears only on the population level;
in each family,
an offspring receives roughly the same quantity
of genetic information from the father and the mother.
The new principle becomes apparent
only if we consider not a family
but the entire population.
(We discuss an "ideal" population,
that crossbreeds randomly.)

\section{A section of the channel for genetic information transmission}
\hspace*{\parindent}
The specialization described above
originates in the fact that
the potential of a male for the transmission of genetic information
is incomparably higher than that of a female.
Every male might, in principle, become the father of the entire progeny
of the population, whereas the ability of females in this respect
is limited.
  
In the jargon of cybernetics, 
the section of the channel
for transmitting male genetic information 
to the progeny
is significantly wider than that 
for transmitting
female genetic information.
Genetically rare males, 
unlike genetically rare females, can play a
substantial role in changing the average genotype,
i.e. the quality of the population.
  
The difference between the two sections of the channel to the progeny
also appears in that each separate male "tends"
to use his ability to the largest degree and leave
a maximum quantity of progeny, thus affecting
the quality of the population.
At the same time,
every separate female
"tends" to assure 
the best possible quality
for her limited quantity of progeny.
  
The schematic formula looks as follows:

\begin{center}
The number of females determines the size of the population.
\\
Each female fights for the quality of her progeny.
\\
The number of males determines the quality of the population.
\\
Each male fights for the number of his progeny.
\end{center}

To a certain degree,
this simplified formula explains
the different psychologies of the sexes.
Darwin gives an incisive example of this difference:
"Males of the deerhound
are inclined to someone else's females,
whereas the females prefer the company of familiar males."
[Reverse translation from Russian.]
This does not mean, of course,
that the females are good and the males are bad.
They are simply different, and the difference
has a biological foundation.
  
To understand the advantages of sexual specialization,
we have to consider the relationship between
the population and the environment.
But first we will try to reveal the features
of analogous mechanisms in various control systems
from the point of view of cybernetics,
the science of analogies.

\section{What do a rocket, a soccer team, and an animal population have in common?}
\hspace*{\parindent}
From the point of view of cybernetics,
all three are control systems.
All are characterized by the attempt to reach a goal.
The goal may be the Moon for the rocket,
victory for the soccer team,
survival for the animal population.
All three systems are subject to perturbations.
The rocket is perturbed by the atmosphere and gravitational fields,
the soccer team is perturbed by the efforts of the opposite team,
and the animal population is perturbed by the environmental factors:
climate, food, predators, parasites, etc.
  
Each system counteracts the perturbations to achieve
stability of its motion.
What is the mechanism of counteraction?
  
An important feature
is separation of the mechanism for conservation,
with the task of keeping everything as is,
from the mechanism for making changes,
with the task of correcting errors.
In the rocket, stabilizers [fin assembly]
are for conservation
while rudders are for making changes in the motion.
In the soccer team, these mechanisms are respectively
the defense players, who try to keep the score
constant, and the forwards, who try to change it 
in favor of their team.
The same result, stability of motion,
is achieved in a similar manner in different systems:
by separate mechanisms for conservation and 
for making changes.
This separation allows the system 
to have maximum stability of motion.
  
What about the animal population?
Does not the differentiation of sexes relate to
the analogous separation of 
conservation and change?
  
We have already shown 
that the females control the quantitative side
of reproduction and the males define the qualitative side.
In the biological categories, this means that the females
largely express the tendency
for inheritance, and the males largely express
the tendency for change.
Finally, in computer science terminology,
the female represents the memory in a permanent storage
for the species,
whereas the male represents the memory in a temporary storage.
  
This separation of two types of memory
gives substantial advantages to the species.
To see this let us consider the relationship
of the population with the environment.

\section{The front between the harmful factor and the curve of mortality}
\hspace*{\parindent}
The notion of environment includes the set of all physical, chemical,
and biological factors which affect an organism during its lifespan.
Climatic factors are part of the environment:
cold and heat, humidity and drought;
these also include various chemicals in the food, water, and air;
and finally there are various animals of the same or different species
which live nearby (predators, parasites, etc).
  
A characteristic feature of a living system
is the ability to adapt to changing environmental conditions.
The system must receive information 
about the change from the environment
in order to adapt.
  
All characteristics of an organism are directly or indirectly connected
with the corresponding environmental factors:
the resistance to cold is connected with the low temperature,
the resistance to heat is connected with the high temperature,
the resistance to drought is connected with the humidity etc.
Connections with other factors of the environment may be less evident;
however, it is beyond doubt that 
the optimal, average values for 
the characteristics of an organism are,
in the end, determined by the corresponding
factors or combinations of factors of the environment.
  
For expository simplicity 
let us choose a factor-characteristic pair,
say temperature and the resistance to temperature,
and represent graphically in Fig.7.1
the relation between the population and the environment
for this particular pair.
  
In Fig.7.1, the abscissa 
represents the intensity of the harmful perturbation factor,
say cold;
at the same time, 
the abscissa represents the degree of resistance to the factor.
The ordinate in Fig.7.1
represents
the number of animals 
that would die
for a particular value of the factor,
i.e. as a result of a given temperature.
The curve obtained characterizes
the mortality of the population as a whole.
The harmful factor front is represented by
a vertical line.
The front cuts off the 
part of the population 
(the shaded area in Fig.7.1)
which is
most vulnerable to the factor.
[More explanation,
might be helpful.
An additional dimension, time, is implied
in Fig.7.1.
The term "harmful factor front" suggests the 
"approaching" factor, i.e. the weather is becoming colder.
The vertical line called the "front" in Fig.7.1,
represents
the current level of the factor in this case.
If, on the other hand, the factor is "retreating",
i.e. the weather is becoming warmer,
then the current level of the factor 
represents its "rear."
Fig.7.1 does not represent the latter case
as
the factor should not to be considered harmful
when it is "retreating":
specimens do not die
because of it.
But they may start dying from 
the opposite factor which then becomes
harmful, i.e. heat.
{\em Translator thanks
Dr. J.B. Kruskal who pointed out this and several
other somewhat non-obvious spots in the original.}]

The mortality curve must be always 
in contact with the harmful factor front
in order for the population to sense the approaching
perturbation.
The population must always sacrifice some quantity
to receive the information which improves the population quality
with respect to the factor.

\begin{figure}
\includegraphics*[width=6.2in]{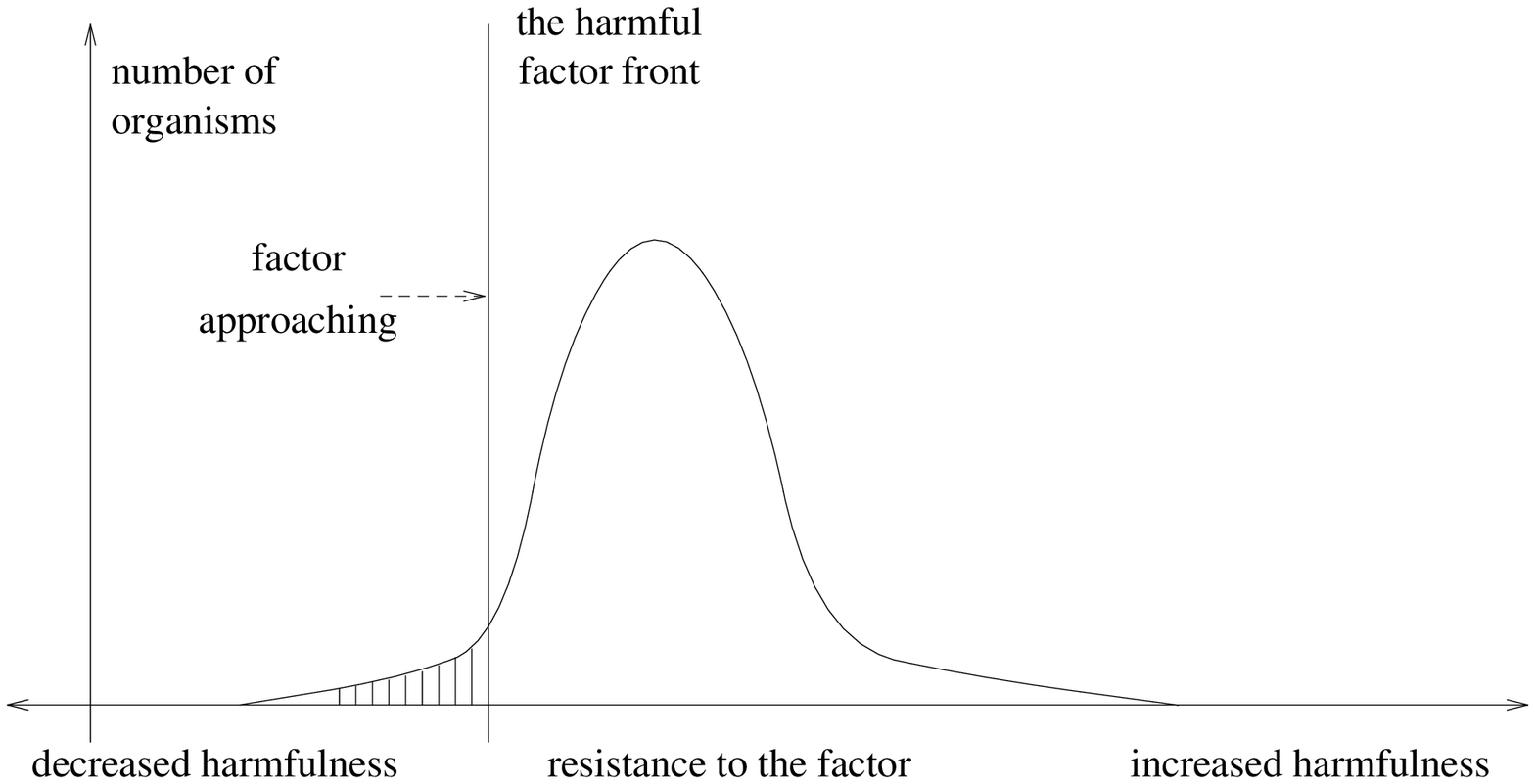}
\caption{
}
\end{figure}

This implies, for example,
that even in a population living in the tropics,
say, monkeys, some animals sometimes die
from the cold.
At the same time, even in an Arctic or Antarctic population,
say, penguins or white bears, 
some animals sometimes die from the heat.
  
A tribute for the information received
is necessary for the informational contact with the environment.
If a population pays no such tribute,
no information is received from the environment
and the population is not able to adapt to it.
A sudden change in the environmental condition
might have caught such a population unprepared and killed it entirely.
  
Naturally, it is advantageous for a population to minimize
the tribute of the quantity for a new quality.
How is this minimization achieved?
  
The optimization method employs the asymmetry
of the relationship of the sexes 
to the quantity and quality of progeny.
Caused by a perturbation in the environment,
losses of males or females 
affect progeny differently.
A loss of females strongly affects 
the quantity of the progeny,
without essentially affecting the quality.
On the other hand, 
a loss of males
during unfavorable environmental conditions
does not tell on the quantity of the population but
changes the quality in the proper direction.
  
Thus, we may say, that a loss of females during "hardships"
is useless, decreasing the size of the population.
A loss of males in similar conditions is useful,
promoting the evolution of the species.

\section{Is the "beautiful" sex really "weak"?}
\hspace*{\parindent}
Poets and writers often call
the female sex "beautiful" and "weak."
The validity of the first attribute seems
undoubted.
But is the second one correct?
  
If by strength we understand
the degree of resistance in hardships,
then the female sex should be considered as the strong one.
Indeed, multiple experiments on plants and animals
and observation on humans show that 
the males die first 
as a result of all harmful
environmental factors:
heat, cold, hunger, poisons, and illnesses.
Not only does the entire male organism
have
a lower resistance than the female organism,
but its various organs, tissues and cells
are also weaker
than those in females.
  
How can 
this weakness and higher mortality of males
be explained?
Two theories explain the phenomenon.
The first one says that the heterogametic sex
always has a higher mortality because of recessive genes
connected with the sex chromosome.
[The heterogametic sex is the one
with unlike sex chromosomes X and Y,
as opposed to the homogametic sex which possesses
similar sex chromosomes X and X.
If a recessive bad gene occurs
in a sex chromosome of a heterogametic specimen,
the latter acquires the bad trait because 
there is no similar gene in the specimen's genetic set.
Hemophilia is a classical example of such a trait.
The gene responsible for the illness is located in the sex chromosome X.
Unlike men,
women carrying the hemophilia gene have a low chance
of acquiring the disease because 
their genetic set includes another X chromosome
which usually masks the sick gene
with a healthy counterpart.
{\em Translator's comment}.]

As to the second theory, it infers a higher male mortality
from a more intensive metabolism.
  
The first theory contradicts the results 
of mortality studies 
among birds, butterflies and moths.
Unlike the overwhelming majority of other species,
the females of these species constitute the heterogametic sex
while the males are the homogametic sex.
Yet experiments show that in many species
of butterflies, several species of birds and moths,
male mortality is almost always 
higher than that of females.
  
The second theory, in fact, does not explain
anything but substitutes
one incomprehensible phenomenon of higher mortality
with
another no more comprehensible phenomenon
of higher metabolism.
If resistant females
exist,
why should not similarly resistant males?

\section{The higher mortality of males is ... expedient}
\hspace*{\parindent}
The fact that the males are biologically weaker
implies the following:
If, 
on the same graph,
we draw the mortality curves for each sex separately,
then we will see that only the males' curve makes contact
with the front of a harmful factor.
There are two main possibilities 
for drawing the females' curve.
The latter can either shift to the right
(Fig.9.1),
or it can have a smaller dispersion 
than the males' curve (Fig.9.2).

\begin{figure}
\includegraphics*[width=6.2in]{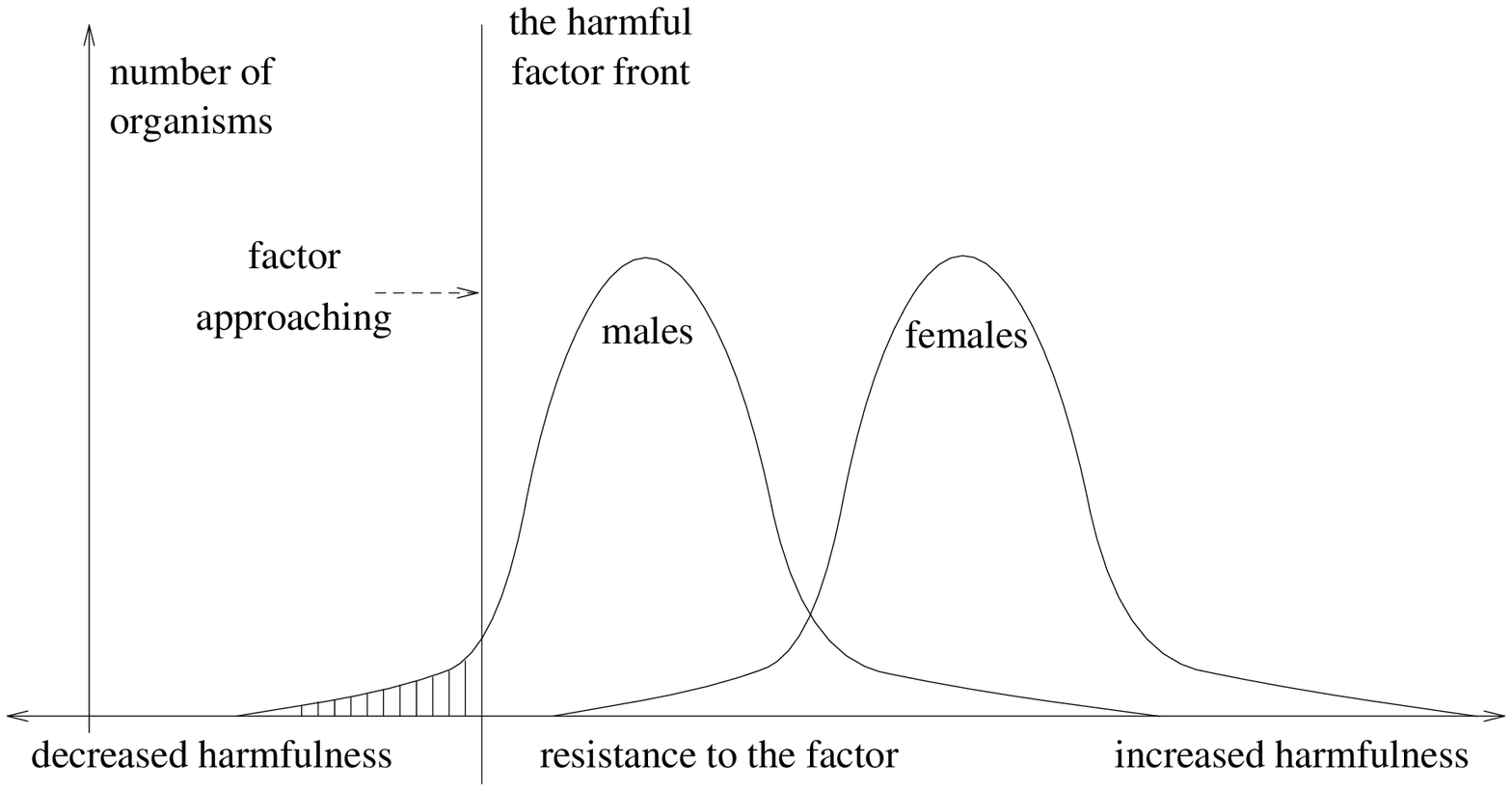}
\caption{
}
\end{figure}

Taking into account our previous considerations
(that the loss of males does not affect the quantity of the population but
promotes adaptation of quality, and that rare specimens of males have
a larger informational value than rare specimens of females;
that the males are the main carriers of the information from the environment
to the population),
we come to the conclusion that 
for each 
characteristic
considered,
the males' curve 
must have larger dispersion than that of the females.
This means that the males must have greater variety
that the females in all qualities.

\begin{figure}
\includegraphics*[width=6.2in]{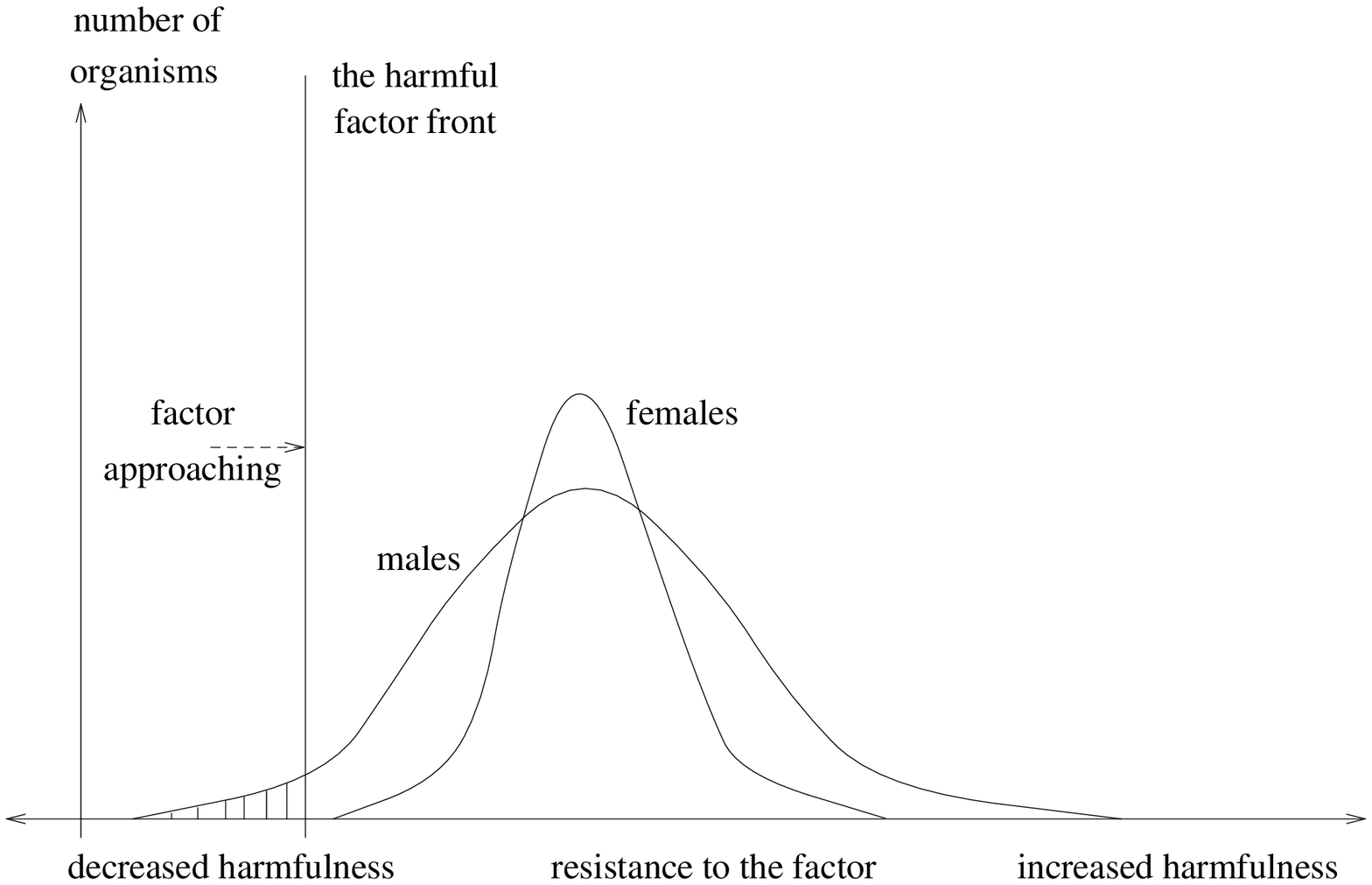}
\caption{
}
\end{figure}

If we 
include all the males in the population in one team
and 
include all the females in the other
and arrange competitions between the two teams,
then the champions 
in all personal competitions will be the males,
whereas in a whole-team competition 
(where the results of all participants count)
the females will be the winners.
[In other words:
the mean-value of the females' curve is to the right of the males'
(females are the winners in a whole-team competition),
while the dispersion of the females' curve is lower than of the males'
(females are the losers in personal competitions).
Thus,
the author suggests that 
a combination of both tendencies takes place,
the one presented in Fig.9.1 and the one presented in Fig.9.2.
{\em Translator's comment}.]
  
Such a relationship of the resistances between the sexes
makes it possible for the population as a whole
to pay for the new information
mainly with males 
whose loss promotes the shift of quality
without changing the number of specimens.
Thus, the higher male mortality is expedient for
the survival of the species.

\section{The sex ratio at birth and the environmental conditions}
\hspace*{\parindent}
It is now clear that the sex ratio is an important parameter of the population,
connected with the hereditary conservation and changeability tendencies
in the reproduction process.
Therefore, the sex ratio at birth 
must reflect
these tendencies in their dependence on the environmental conditions
during different periods
of the population history.
  
An increase of male mortality in unfavorable environmental conditions
must lead to an
increase of the male/female ratio at birth.
An increase of this ratio may be triggered
directly by a changed environmental conditions,
independently of the ratio of sexes of adult animals.
There are reasons to believe that in vertebrates
this increase is controlled
by steroid hormones of the hypophysis (pituitary gland),
adrenal cortex and gonads.
  
To promote faster adaptation of the population,
an environmental hardship should increase
the male's renewal rate.
The rule is that in response to {\em any} unfavorable environmental
condition, {\em both} the birth and the death rates for the males increase.
  
Multiple facts reported and known in biology confirm this rule.
Interestingly enough, the species which utilize either
method of reproduction, asexual or sexual
(bacteria, infusoria, some crayfish, and other),
resort to sexual reproduction under unfavorable
environmental conditions.
  
For example, among many kinds of water-fleas {\em daphnia genus}
as well as among the aphids {\em aphidae},
when the conditions are favorable,
usually in summer, asexual reproduction
(parthenogenesis) takes place.
The new fleas (only females) hatch out of
the summer soft-membrane eggs.
When less favorable conditions strike,
some of the females produce a quantity of males,
which then copulate with the females.
The fertilized females deposit 
the hard-shelled winter eggs, 
which can stay alive
for a long time in unfavorable conditions:
during cold, heat, drought, 
when the water reservoir dries out.
  
Extracting a population of female rotifers {\em rotifera phylum}
from pond water and placing them into river 
or well water, or doing the reverse resettlement,
the scientists observed an emergence of males
on the third or fourth day after the move.
The direction of the move,
from the pond to the river or from the river to the pond,
was unimportant:
any change in the environment caused the males to appear.
  
Subjecting 
sexual animals
like drosophila (fruit flies)
to harmful factors,
the researchers observed
the simultaneous increases in both the male birth and death rates.
This seemed paradoxical and unexplainable.
Indeed, why should a harmful factor
(no matter which:
hunger, cold, heat, or a poison)
act as such for male flies 
at all stages of the life causing them to die more intensively,
except
at the very beginning of the male life
and, moreover, promote the birth of
more males?
Why are the sex theories sometimes so contradictory?
For example, paying attention to the fact that during a cold
year
more boys than girls are born,
a scientist inferred that cold promotes the birth of boys,
while heat promotes the birth of girls.
Later, the scientist noticed that extreme heat also promotes
the birth of the boys.
Then another theory appeared which explained everything
in exactly the opposite way:
heat promotes the birth of boys, while cold
promotes the birth of girls.
  
Meanwhile, both phenomena are explainable easily
by the rule of higher male renewal rate
for any change in the environmental
conditions.
  
Facts which confirm this rule can be found among mammals,
humans including.
Medical and demographic statistics show that
during substantial climatic and social shifts
(abrupt change of the temperature,
drought, war, hunger, resettlement),
that is, during an increase of mortality,
a tendency to increase the ratio of boys to girls
among the newborn babies is also observed.
The same tendency is observed by cattle-breeders:
better the maintenance conditions of the animals,
more females in the progeny,
even if artificial insemination is practiced and the sperm
is taken from the same male.
  
The separation of the population into two sexes
and the specialization of the sexes,
wherein one sex is responsible
for quality and the other
for quantity
leads to the situation where any information stream
about environmental changes is first 
received by the males,
which react to this information and transform
the stream.
In other words, new information
gets first into the temporary 
memory of the population,
where it is checked and selected,
and only after this
is it transferred into the permanent memory,
i.e. females.
  
This separation, into a more inertial stable kernel
and a more mobile sensitive shell, allows the population
to distinguish temporary, short-term and random factors
of the environment, e.g. an unusually cold winter,
or an especially hot summer,
from systematic changes in the same direction,
say the beginning of an Ice Age.
One may say that the information stream from the temporary
memory gets to the permanent memory through
a frequency filter.
The filter lets low frequency through
but blocks high frequencies.
It is this filtration through the temporary memory
by which
the inertiality of the permanent memory
is achieved.

\section{Clay and marble}
\hspace*{\parindent}
A good sculptor,
before making a sculpture out of marble,
will create many models out of clay.
The nature acts similarly.
Like a sculptor, it
first
creates a large variety of males (clay models),
testing them and selecting good versions
to implement later in females (marble sculpture).
Thus, in a population,
new qualities first appear among the males
and may afterward appear among the females.
  
We may consider the male as the vanguard
of the population, which advances
to meet the harmful factors of the environment.
A certain distance is kept between
this vanguard and the kernel,
"the golden fund" of the population.
The distance is necessary for testing and selection.
  
The evolutionary inertiality, lagging of the females,
is a payment for their perfection.
Vice versa,
the progressivity of the males is a benefit of
their imperfection.
  
We can formulate the following hypothesis:
a new quality in phylogeny
must first become permanent among the males,
and then it must be transferred to the females.
In other words, the males are the "door"
for the change in the heredity of the population.
  
Thus, if the male and the female are distinct from each other
in some quality, say in the height or color,
one may predict the direction of change,
namely, the quality is changing in the direction
from the female to the male.
For example,
if males are bigger than the females,
then there is an evolutionary
tendency for size to increase in the species.
In the other case, if males are smaller than females,
the species evolves to have smaller specimens.
  
We may conjecture that humans are becoming taller
at this stage of history,
because an average man is taller than an average woman.
Among the spiders,
the tendency must be opposite, because their males
are smaller than the females.
The anthropologists and entomologists believe
this is the case:
mankind is growing, spiders are shrinking.
  
Another example is the well-known connection
between ontogenetic and phylogenetic
emergence of the antlers
in male and female deers.
A strong relation exists between
the extent of antlers in a species
and the age when antlers appear in a specimen.
Namely,
the larger the extent,
the earlier the antlers appear,
first in males, then in females.
  
The suggested rule can be applied to study some concrete
problem of evolution,
remembering of course that this general tendency
can sometimes be overlapped
by other tendencies.

\section{Conclusion}
\hspace*{\parindent}
An application of certain general ideas and approaches
of cybernetics to the formulation and solution
of biological problems allows us to understand
certain facts, previously mysterious.
Now we know that the advantages of asexual reproduction
are efficient only in short "sprinter" evolutionary distances.
"Stayer" and "marathon" distances need
the sexual method.
The advantages of crossbreeding and differentiation
are now clear, as well as the fact that these advantages
can be fully realized only in an "ideal" population.
This implies an insignificant sexual dimorphism in
monogamous species versus
a strong sexual dimorphism in polygamous species.
  
Returning to our first question, whether males are generally needed,
we should answer: yes, they are needed
mostly for adaptation to the changes in environmental conditions.
This holds for animals. 
What about humans?
It is known that social and technological progress
steadily decreases the role of the biological evolution.
Having learned how to change the environmental conditions,
man renders himself free from the necessity to change himself.
Indeed, if a new Ice Age begins,
animals will grow thick hair,
but man will put on synthetic fur clothes.
  
We conclude that social and technological progress
should steadily increase the role and proportion of women in the society.
[This statement which is apparently saying that
men are dying out is put in such an indirect way 
perhaps to 
improve the chances of the original publication.
{\em Translator's comment}.]
\end{document}